%% file: paper.tex
\documentclass[conference]{IEEEtran}

\usepackage{cite}
\usepackage{comment}
\usepackage{enumitem}
\usepackage{amsmath,amssymb,amsfonts}
\usepackage{algorithmic}
\usepackage{graphicx}
\usepackage{float}
\usepackage{textcomp}
\usepackage{xcolor}
\usepackage{booktabs}
\usepackage{multirow}
\usepackage[nolist]{acronym}
\usepackage{booktabs}
\usepackage{tcolorbox}
\usepackage{listings}
\usepackage[linesnumbered,ruled,vlined]{algorithm2e}
\usepackage{hyperref}
\usepackage{tabularx}
\usepackage{caption} 
\usepackage{mdframed}

\def\BibTeX{{\rm B\kern-.05em{\sc i\kern-.025em b}\kern-.08em
    T\kern-.1667em\lower.7ex\hbox{E}\kern-.125emX}}
    
\begin{document}

\bstctlcite{IEEEexample:BSTcontrol}

\title{Leveraging LLM Agents and Digital Twins for Fault Handling in Process Plants} 

\author{
    \IEEEauthorblockN{
        Milapji Singh Gill\IEEEauthorrefmark{1},
        Javal Vyas\IEEEauthorrefmark{2},
        Artan Markaj\IEEEauthorrefmark{1},
        Felix Gehlhoff\IEEEauthorrefmark{1},
        Mehmet Mercangöz\IEEEauthorrefmark{2}%
    } 
    
    \IEEEauthorblockA{
        \IEEEauthorrefmark{1}Institute of Automation Technology\\
        \textit{Helmut Schmidt University Hamburg, Germany}\\
        {\tt\small \{milapji.gill, artan.markaj, felix.gehlhoff\}@hsu-hh.de}\\
        \IEEEauthorrefmark{2}Autonomous Industrial Systems Lab,\\
        \textit{Imperial College London, United Kingdom}\\
        {\tt\small \{j.vyas24, m.mercangoz\}@imperial.ac.uk}\\
       }
       \vspace{-1cm}
}
\maketitle


\begin{abstract}

Advances in Automation and Artificial Intelligence continue to enhance the autonomy of process plants in handling various operational scenarios. However, certain tasks, such as fault handling, remain challenging, as they rely heavily on human expertise. This highlights the need for systematic, knowledge-based methods.
To address this gap, we propose a methodological framework that integrates Large Language Model (LLM) agents with a Digital Twin environment. The LLM agents continuously interpret system states and initiate control actions, including responses to unexpected faults, with the goal of returning the system to normal operation. In this context, the Digital Twin acts both as a structured repository of plant-specific engineering knowledge for agent prompting and as a simulation platform for the systematic validation and verification of the generated corrective control actions. The evaluation using a mixing module of a process plant demonstrates that the proposed framework is capable not only of autonomously controlling the mixing module, but also of generating effective corrective actions to mitigate a pipe clogging with only a few reprompts.
\end{abstract}

\input{01_Introduction}

\input{02_Background}

\input{03_State_of_Art}

\input{04_Requirements}

\input{05_Methodology}

\input{06_Exemplary_Application}

\input{07_Discussion}

\input{08_Summary_and_Outlook}

\section*{Acknowledgment}

This research [project ProMoDi] is funded by dtec.bw – Digitalization and Technology Research Center of the Bundeswehr. dtec.bw is funded by the European Union – NextGenerationEU.
Financial support from ABB for the Autonomous Industrial Systems Laboratory at Imperial College London is gratefully acknowledged. The first two authors contributed equally to this work.

\input{references.bbl}

\end{document}

%% file: 01_Introduction.tex
\section{Introduction} \label{intro}

Modern automation systems have streamlined many routine operations in industrial environments, but fault handling remains a cognitively demanding and predominantly manual process. Experienced operators are required to react immediately to anomalous behavior by selecting appropriate corrective control actions \cite{Markaj.2024}. These tasks are typically highly situational, difficult to generalize, and often performed under time pressure. In complex technical systems such as process plants, the same observable symptom may stem from multiple root causes, each requiring a different response \cite{Webert.2022, Manca.2021, Westermann.2023}. This ambiguity is rarely captured in predefined operator instructions or static fault handling strategies, making human expertise indispensable \cite{Markaj.2024}. As a result, fault handling is not only labor-intensive but also prone to error and, in some cases, safety-critical. 
These challenges, combined with increasing plant complexity and the demographic shift in the workforce leading to a shortage of experienced operators, highlight the urgent need for more autonomous 
solutions \cite{Manca.2021, Markaj.2024}.

To address these limitations, recent research has turned to Artificial Intelligence (AI) methods \cite{Gill.2024, Liu.2023, Manca.2021}. Machine Learning (ML) is effective for detecting anomalies as deviations from expected behavior \cite{Westermann.2023} but generally lacks the capability to suggest concrete, executable responses for novel fault types. Large Language Models (LLMs), by contrast, have garnered considerable attention due to their advanced reasoning and generalization capabilities. Unlike conventional ML models, LLMs offer a versatile reasoning mechanism that makes them adaptable to various industrial control applications \cite{GPT4IAS,Vyas24}. However, several challenges remain with regard to fault handling within process plants. LLMs often lack plant-specific knowledge \cite{Lewis.22.05.2020}, which can be extracted from systems engineering artifacts \cite{Hildebrandt.2017}, leading to hallucinations and unsafe plant states. Moreover, fault handling typically requires sequential reasoning steps \cite{Piardi.2025}. Executing these steps autonomously and reliably, particularly in response to unknown fault types, requires more than isolated AI components. It needs structured orchestration of perception, reasoning, and action. Thus, to effectively develop and deploy LLM-based fault handling solutions in technical systems, a reusable methodological framework is essential. This leads to the following open Research Questions (RQs):
\begin{itemize}
    \item \textbf{RQ1:} \textit{How can a methodological framework be designed that enables LLMs to handle unknown fault types in process plants, while reliably ensuring the operational safety of their proposed corrective actions?}
    \item \textbf{RQ2:} \textit{Can systems engineering information help LLMs generate and execute effective corrective actions, and how should it be represented in the prompt?}
\end{itemize}    
The remainder of this paper is structured as follows: Sec.\ref{background} provides background on fault handling in process plants and recent advances in LLM technologies, motivating their potential use for autonomous fault handling. Sec.\ref{Related Work} reviews related work on LLM-based plant control. Based on these insights, Sec.\ref{Requirements} derives requirements for the proposed framework. Sec.\ref{Autonomous Fault Handling} introduces the framework, including a prompt engineering approach. Sec.\ref{exemplary application} presents the experimental setup and evaluation results using a mixing module. Key findings are discussed in Sec.\ref{discussion}, and Sec.\ref{summary and outlook} concludes the paper with an outlook on future research.

%% file: 02_Background.tex
\section{Background} \label{background}
\subsection{Fault Handling in Process Plants}

In modern process plants, fault handling relies on continuous monitoring of process parameters via control systems, dashboards, and alarm mechanisms. Deviations from normal operation trigger alarms, prompting operators to assess the situation and determine corrective control actions \cite{Manca.2021,Markaj.2024}. This assessment draws on knowledge of causal dependencies between process variables \cite{Abele.2013} and requires interpreting real-time data in light of historical trends and plant-specific experience \cite{Manca.2021, Thambirajah.2009}. Based on this, operators initiate corrective actions to stabilize the system or transition it to a safe state, often manually or via control logic \cite{Markaj.2024}. 

To support this task, operators use a range of systems engineering artifacts, including piping and instrumentation diagrams (P\&ID), state machines, control logic, procedures, simulation models, and alarm logs \cite{Thambirajah.2009, Kirchhubel.2020}. These span three semantic layers: structural (component topology), functional (material, energy and signal flow), and behavioral (system dynamics) models \cite{Westermann.2023, Manca.2021}. The Digital Twin concept has been explored to consolidate this plant-specific knowledge, encompassing data, digital models, and digital services, into a structured representation of the physical system to enhance decision support \cite{Tao.2019, Gill.2024}.

\subsection{Large Language Models}

Recently, LLMs have gained a lot of attention for their advanced reasoning capabilities, making them suitable for complex decision-making across diverse contexts \cite{Vyas24, GPT4IAS}. Generally, LLMs are pre-trained transformer-based architectures that predict the next token in a sequence based on patterns learned from massive textual datasets \cite{AshishVaswani.}. LLMs operate purely on linguistic patterns without direct grounding in physical environments \cite{Liang.}. Although they do not possess an internal model of the environment, LLMs can infer plausible continuations or conclusions from textual input \cite{Bubeck.22.03.2023}. 

Despite this capability, it is advisable to explicitly encode task-specific information in the prompt, positioning prompt engineering as a critical interface between domain knowledge and model behavior. Depending on task complexity, domain specificity, and reliability requirements, techniques such as zero-/few-shot prompting, chain-of-thought reasoning, structured templates, and instruction tuning are used to guide model outputs \cite{Vyas24, Xia.}. In this  context, mechanisms like Retrieval-Augmented Generation (RAG) can dynamically supplement prompts, though latency and complexity may limit their use in real-time or safety-critical applications \cite{Lewis.22.05.2020}. Thus, carefully balancing the amount and relevance of information is crucial when using LLMs. Insufficient context increases the risk of hallucinations, while unstructured input may impair the model’s capacity to extract pertinent information \cite{Bubeck.22.03.2023}. 

While LLMs alone remain passive language processors, recent advances in LLM-based agents integrate LLMs with external tools, memory, APIs, and planning mechanisms to enable iterative problem-solving and goal-directed behavior \cite{GPT4IAS}. Unlike static prompting, agent-based architectures allow active task decomposition, structured data retrieval, and function execution, supporting more complex workflows such as control of technical systems \cite{Vyas24, GPT4IAS}. Given these capabilities, LLM agents appear particularly promising for autonomous fault handling, and thus form the focus of our approach described in the following.

%% file: 03_State_of_Art.tex
\section{Related Work} \label{Related Work}

Recent studies investigate the integration of LLMs into industrial control applications by developing frameworks that embed LLMs into automation and control workflows.

In Heating, Ventilation, and Air Conditioning (HVAC) systems, for instance, LLMs have been applied to control tasks, achieving comparable or even superior performance to Reinforcement Learning (RL)-based approaches \cite{HVACLLM}. In parallel, other researchers have focused on the generation of Programmable Logic Controller (PLC) code using LLMs. Through iterative user-guided pipelines and external verification tools, limitations of traditional PLC programming have been addressed. This work culminated in the development of the LLM4PLC package \cite{LLM4PLC}. Building upon this, the Agent4PLC framework introduced a multi-agent architecture powered by LLMs and extended with code-level verification, chain-of-thought prompting, and RAG techniques to support more robust industrial control scenarios \cite{Agent4PLC}.

Additional frameworks for modular and batch production processes have demonstrated how LLM agents can coordinate sequences of atomic control functions to accomplish complex tasks \cite{GPT4IAS}. In this context, end-to-end automation have also been proposed, embedding LLMs into industrial control pipelines for broader system management tasks. A representative example by Xia et al. \cite{GPT4IAS} introduces LLM-based agents for orchestrating modular production processes. Here, LLM agents are embedded within Digital Twin environments and automation systems to plan and control operations based on structured instructions. Modular control is realized via Asset Administration Shells and REST interfaces. Such agent-based systems enable greater flexibility and adaptability in modular production environments.  While the framework highlights the orchestration capabilities of LLMs, it does not include mechanisms for detecting or responding to faults in operation. In a separate line of work, Xia et al. \cite{Xia.} also propose the use of LLM agents to enhance Failure Mode and Effects Analysis (FMEA) for risk management in technical systems. This approach uses a multi-agent architecture and RAG methods to enrich traditional FMEA tables with domain knowledge. Although this supports systematic documentation and identification of risks, the method is limited to static analysis and does not include operation.

In conclusion, despite recent advances, current research has primarily focused on the planning and orchestration of processes, as well as static analysis of process plant-related textual documents. This highlights the need for concepts that enable safe and adaptive fault handling with LLM agents.

%% file: 04_Requirements.tex
\section{Requirements} \label{Requirements}

Drawing on insights from Sec.\ref{background} and Sec.\ref{Related Work}, we define requirements \textbf{(R)} for the methodological framework:

\textbf{(R1) Distributed Task Allocation}: The methodological framework must enable the decomposition of the overall fault handling process into distinct, interacting components solving specific sub-tasks (e.g. monitoring, fault detection, as well as control and corrective control actions) independently \cite{Piardi.2025}. Simultaneously, collaboration is necessary to ensure coherent decision-making across these sub-tasks. This modularization reflects the inherently distributed nature of fault handling in technical systems \cite{Cerrada.2007}.

\textbf{(R2) Adaptive Fault Handling Reasoning Capabilities:} The methodological framework must incorporate intelligent components capable of adaptive reasoning and inferencing to autonomously derive, adjust, and justify corrective control actions in response to previously unknown or uncertain fault scenarios \cite{Piardi.2025}. Such capabilities are essential in fault handling, where rigid rule-based systems often fail to address novel or context-specific issues.

\textbf{(R3) Closed-Loop Action Verification and Validation:} The methodological framework must support automated verification and validation mechanisms of proposed corrective control actions to ensure safe and reliable process execution. This includes monitoring the effects of actions and iteratively refining them when unwanted behavior is detected \cite{Piardi.2025, Vyas24}. Additionally, this loop must account for a maximum allowable time window within which valid corrective control actions must be identified, in order to minimize latency between fault detection and implementation. If no valid solution can be derived within this time frame, manual intervention or predefined safety mechanisms must be triggered \cite{Vyas24}.

\textbf{(R4) Inclusion of Domain Knowledge:} The methodological framework must support the integration of domain-specific system knowledge into the reasoning process, which is relevant to the fault context \cite{Manca.2021, Westermann.2023}. Since LLMs lack direct grounding in physical systems \cite{Liang.}, structured domain input is essential to enable valid inference and reliable fault handling.

\textbf{(R5) Transparent and Traceable Decision-Making:} The methodological framework must ensure transparency and traceability of decision-making processes, enabling human operators to understand how and why certain corrective control actions were proposed or executed. This is essential not only for system validation and continuous improvement but also for enabling human intervention in safety-critical situations \cite{Cummins.15.01.2024}.

%% file: 05_Methodology.tex
\section{Methodological Framework for Autonomous Fault handling in Process Plants} \label{Autonomous Fault Handling}

\subsection{Framework}
\begin{figure*}
\centering
\includegraphics[width=0.9\textwidth]{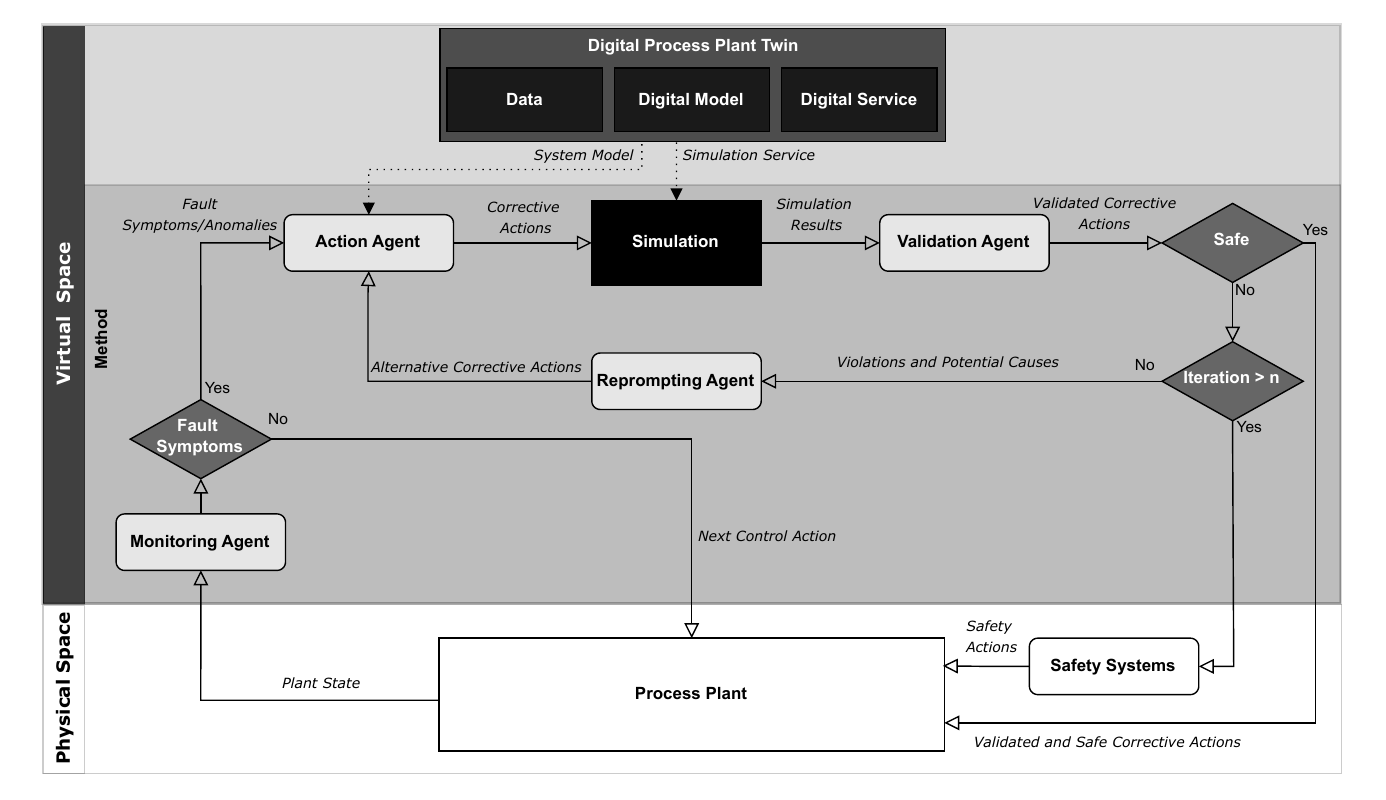}
\caption{Agent-based methodological framework for autonomous fault handling.}
\label{fig:AgenticFramework}
\end{figure*}

In the following, we present a methodological framework that integrates a \textit{Digital Process Plant Twin} and a structured method for orchestrating \textit{LLM-based agents} to enhance the autonomy of fault handling in industrial process plants. An overview of the proposed framework is shown in Fig. \ref{fig:AgenticFramework}. Following the common structure of Cyber-Physical Systems (CPSs), the architecture is divided into a physical and a virtual space. The physical space comprises the real-world \textit{Process Plant}, while the virtual space hosts different components of the methodological framework. Both spaces are closely interconnected and exchange information continuously. The physical \textit{Process Plant} consists of various interconnected physical 
components. In addition to the piping system, key elements include valves, pumps, mixers, and tanks that interact to perform the desired operations. 
The information flow within the method is represented by solid lines, indicating the direct information flow between agents. In contrast, dashed lines indicate the use of the \textit{Digital Process Plant Twin} within the method. 

For the transition from manual to autonomous fault handling, the structured, feedback-driven method iteratively refines agent responses to reduce human intervention while maintaining operational safety. To achieve this, the framework distributes operator responsibilities across distinct interacting agents that reflect common cognitive capabilities in fault handling (see \textbf{R1}). The method incorporates a \textit{Monitoring Agent}, an \textit{Action Agent}, a \textit{Validation Agent}, and a \textit{Reprompting Agent}. These agents incorporate AI methods, when cognitive capabilities are required. In particular, the use of LLMs is motivated by their demonstrated capability to generalize from examples and infer plausible corrective control actions from structured context, as discussed in Sec.~\ref{background} and specified in \textbf{R2}. This makes them particularly well suited for supporting fault handling in systems with incomplete data or ambiguous fault symptoms. The core interaction between agents follows a closed-loop structure that ensures traceability, validity, and verifiability of generated corrective control actions (see \textbf{R3}). Within the method, the agents can access plant-specific knowledge encapsulated in the Digital Twin’s data, digital models, and services (see \textbf{R4}). Additionally, they are also capable of feeding back new data and insights. This tight integration ensures that the agents operate on an up-to-date representation of the \textit{Process Plant} while continuously enriching the Digital Twin’s knowledge base.  To ensure transparency and traceability of decision-making processes, the framework employs a chain-of-thought prompting strategy (see \textbf{R5}). This method enables LLM agents to explicitly articulate their reasoning for each corrective control action, thereby supporting post-hoc interpretation, validation, and human oversight.

The method starts in the virtual space with the \textit{Monitoring Agent}. This agent observes the current state of the physical plant using sensor data, performance monitors, alarm indicators, and diagnostic thresholds. It identifies potential fault symptoms, based on deviations from nominal behavior. If no fault symptoms are detected, the next control action is executed to maintain normal process operation. If fault symptoms are identified, the \textit{Action Agent} leverages an LLM-driven approach to synthesize corrective control actions based on the current system state. By consulting plant-specific information from the \textit{Digital Process Plant Twin}, previous interactions, and alternative actions, the agent generates a set of potential helpful corrective control actions to mitigate the fault. These proposed actions are tested in the \textit{Simulation}, which is accessed as a service from the \textit{Digital Process Plant Twin} to assess their impact and potential unintended consequences.  This \textit{Simulation}, as a virtual replica of the \textit{Process Plant},  provides a risk-free setting where generated corrective control actions can be validated and verified without exposing the plant to additional hazards. Moreover, it allows for the fine-tuning of corrective control actions under a variety of simulated fault conditions. Post-simulation, the \textit{Validation Agent} plays a crucial role in assessing the feasibility, safety, and overall effectiveness of the corrective control actions. It ensures that any action proposed for real-world deployment adheres strictly to operational protocols and safety standards. For validation purposes, various methods can be applied. Among others, a cost function incorporating multiple influencing factors, such as process stability, energy consumption or control effort, can be included to assess the suitability of the generated actions. In cases where the initial corrective control action fails to meet the required validation criteria, the \textit{Reprompting Agent} intervenes. This agent iteratively refines the corrective strategy by incorporating feedback from the \textit{Simulation}. Through successive iterations, the \textit{Reprompting Agent} optimizes the proposed response until a valid solution is identified. Once an corrective control action passes validation (see Fig. \ref{fig:AgenticFramework}, decision point \textit{Safe}), the corrective control action is passed to the \textit{Process Plant}. The \textit{Safety System} acts as a fallback mechanism when no valid corrective control action can be found after a defined number of iterations. These systems implement pre-defined emergency protocols, including shutdown procedures or manual intervention.

In this contribution, we specifically focus on the iterative loop between the \textit{Action Agent}, the \textit{Simulation}, the \textit{Validator Agent}, and the \textit{Reprompting Agent}, as these execute the essential method steps of the proposed framework. To enable the derivation of corrective control actions, relevant information about the \textit{Process Plant} must be made available to the LLMs. The following subsection V-B details the prompt engineering strategy employed for the \textit{Action Agent} and the \textit{Reprompting Agent}, both of which are critical components within the loop.

\subsection{Prompt Engineering in LLM Agents using Digital Twin Information} \label{Prompt}

To enable the effective and efficient generation of corrective control actions, we designed a prompt structure that supplies both LLM agents (\textit{Action Agent} and \textit{Reprompting Agent}) with task and plant-specific knowledge. 

The prompt is generally structured into the three main sections: \texttt{<Agent Description>}, \texttt{<Plant Description>}, and \texttt{<Agent Action>}. Each of these overarching sections comprise more specific subsections that provide detailed contextual information to support the LLM’s reasoning and decision-making. An excerpt of the prompt used for the \textit{Action Agent }is illustrated in Fig. \ref{fig:prompt-structure}. Looking at the \texttt{<Agent Description>}, the agent receives a \texttt{[Role]}, outlining its responsibilities. The main task is then described in terms of the \texttt{[Goal]} and the \texttt{[Task]}. The section \texttt{<Agent Action>} specifies the \texttt{[Expected Output]} from the LLM, which is then processed by the scripts described in Sec.~\ref{exemplary application} to re-execute, validate, and verify the proposed corrective control actions within the loop. The essential part of the prompt in our approach is the \texttt{<Plant Description>} section which provides plant-specific information. It details the \texttt{[Plant Function]}, the \texttt{[Plant Structure]} as well as the intended process sequence in \texttt{[Plant Behavior]}. To ensure situational awareness, the prompt dynamically integrates the \texttt{[Current Plant State]}. This prompt structure aligns with systems engineering principles, where a system is conceptually described in terms of \textit{structure}, \textit{function}, and \textit{behavior} \cite{Hildebrandt.2017}. Structural aspects can be derived from engineering artifacts such as P\&IDs. Functional roles describe how each component contributes to the process goal, while behavioral logic is encoded using formal models such as Finite-State Machines to represent state transitions and causal dependencies. This supports the LLM agent’s understanding of permissible actions and transition conditions. 

An essential feature of the prompt design is that the \texttt{<Plant Description>} is treated as a variable input in terms of how formal the information is described. This approach enables the use of heterogeneous modeling representation formats, which typically exist in a plants lifecycle. 

Supported input formats range from informal text-based specifications and semi-formal models like SysML class diagrams to formal representations such as simulation code or domain-specific ontologies \cite{GPT4IAS,Gill.2024}. While the LLM does not operate directly on these models, their contents are converted into structured natural language or graph-to-text renderings for prompt integration. This design choice decouples the prompt format from the underlying modeling formalism, thus enhancing the generalizability and extensibility of the architecture across different domains and abstraction levels. Embedding this information directly into the prompt ensures that the LLM agent has consistent and complete access to the relevant context at inference time, without incurring latency or inconsistencies introduced by runtime retrieval.

\begin{figure}[!ht]
\centering
\begin{tcolorbox}[title=, colback=gray!5, colframe=gray!80!black, fonttitle=\ttfamily\bfseries, width=\linewidth]
\begin{lstlisting}[language=, basicstyle=\ttfamily\tiny]
<Agent Description>

[Role]
Plant operator: Ensures safe operations of the chemical plant.

[Goal]
Maintain plant operation within safety limits and execute corrective actions 
if deviations occur.

[Task]
- Sequentially fill and empty tanks B201 to B204.
- ...

[Skills]
- Ensure safe plant operation
- ...
------------------------------------------------------------------------
<Plant Description>

[Plant Function]
- Mixing of three liquids, sequentially transferred from tanks B201,  B202, and B203 into tank B204.

[Plant Structure]
- The system consists of four tanks: B201, B202, B203, B204
- There are eight valves controlling the liquid movement:
    - Filling Valves: valve_in0, valve_in1,valve_in2
    - ...

[Plant Behavior]
- Control sequence (step 1 to 9)

[Current Plant State]
- tank_B201_level: 0.020m
...
------------------------------------------------------------------------
<Agent Action>

[Expected Output]
- Operation Action list for operation (e.g., "valve_in0 - close")
...
\end{lstlisting}
\end{tcolorbox}
\caption{Prompt structure with exemplary natural text information provided to the \textit{Action Agent} and \textit{Reprompting Action}.}
\label{fig:prompt-structure}
\end{figure}

%% file: 06_Exemplary_Application.tex
\section{Evaluation} \label{exemplary application}

\subsection{Experimental Set Up and Implementation}

\begin{figure}
    \centering
    \includegraphics[width=1\linewidth]{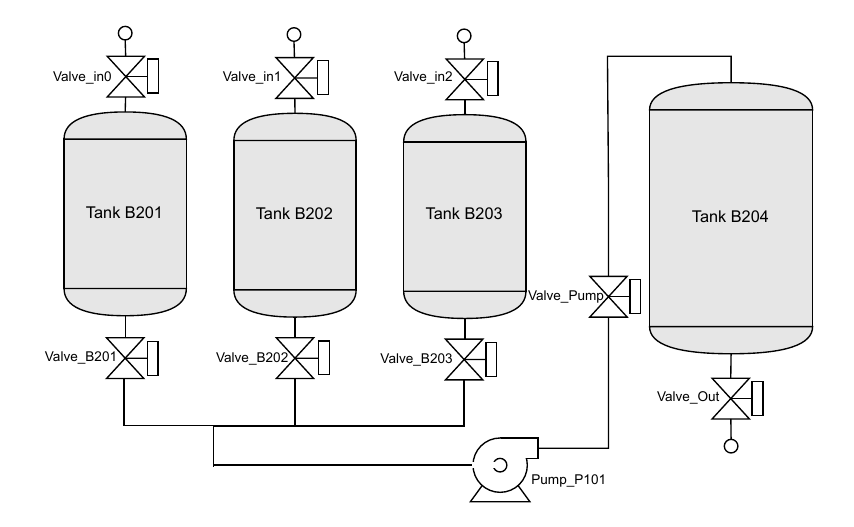}
    \caption{Mixing module with all relevant actuators}
    \label{fig:mixing module}
\end{figure}

We base our experimental set up on a benchmark introduced by Ehrhardt et al. \cite{J.Ehrhardt.2022}, designed to evaluate AI-based diagnosis, reconfiguration, and planning in a modular Process Plant. The provided simulation model within this benchmark, which describes a mixing module, incorporates parametrized fault types, including clogging, leakage, and pump degradation, making it suitable for evaluating the proposed methodological framework. The mixing module, depicted in Fig. \ref{fig:mixing module}, is implemented in \texttt{Open Modelica}. It models a four-tank system (\texttt{tank\_B201} - \texttt{tank\_B204}) with a central pump (\texttt{pump\_P101}) and controllable valves (e.g. \texttt{valve\_in0}). Liquid is filled into \texttt{tank\_B201} - \texttt{tank\_B203} and sequentially transferred to \texttt{tank\_B204}. State transitions are managed via discrete logic blocks, with condition monitoring based on level sensors (e.g., \texttt{sensor\_discrete\_tank\_B203\_high}), pressure sensors (e.g., \texttt{sensor\_continuous\_pressure
\_tank\_B202}), and volume flow rates (\texttt{ sensor\_continuous\_volumeFlowRate}). As a test case, we focus on a clogging fault scenario, which requires multi-step reasoning. In this setting, the LLM must first detect the anomalous condition independently, based on sensor values and subsequently respond by increasing the power of \texttt{pump\_P101}. Given the number of potential control options, this scenario presents a non-trivial fault condition for evaluating both the reliability and efficiency of generated actions based on the actual number of reprompts needed.
 
To implement the proposed methodological framework, we use a modular orchestration implemented in \texttt{Python}. For the purpose of this case study, the \textit{Action Agent} is implemented as the \texttt{PlantOperatorCrew}, the \textit{Validation Agent} as the \texttt{validation\_script.py}, and the \textit{Reprompting Agent} as the \texttt{PlantStrategyCrew}. The script \texttt{main.py} coordinates the iterative interaction between (i) a simulated plant model, (ii) a Digital Twin, and (iii) LLM-based agents implemented using \texttt{CrewAI}. Initial conditions, actuator states, and fault parameters are passed via a dictionary-based configuration. At each iteration, current plant states are passed to the \texttt{PlantOperatorCrew}, which uses an LLM, in our case \texttt{GPT-4o} or \texttt{GPT-4o mini}, to propose corrective control actions based on a structured prompt format. This format, as introduced in Sec.~\ref{Prompt}, is operationalized through two \texttt{YAML} files (\texttt{agents.yaml} and \texttt{task.yaml}) (see Fig.~\ref{fig:prompt-structure}). The plant states, along with fault configurations and simulation parameters, are defined in a centralized \texttt{JSON} configuration file, which serves as a persistent interface between modules. 

The control actions proposed by the agent are written back into this \texttt{JSON} file and applied to the \textit{Digital Twin Simulation}. If these actions are deemed valid by specified rules in the \texttt{validation\_script.py}, the simulated plant model is updated accordingly. Otherwise, the \texttt{PlantStrategyCrew} generates improved suggestions based on the flagged issues. This loop continues until a stop condition is reached (i.e., \texttt{tank\_B204} reaches the target level). Plant states, control actions, and token usage are logged at each step. \texttt{CSV} exports (\texttt{plant\_op.csv}, \texttt{digital\_twin\_op.csv}, and \texttt{llm\_plant\_op.csv}) enable further analysis of both plant performance and prompt efficiency. A pseudocode summary of this control loop is shown in Algorithm \ref{alg:llm-loop}. To minimize variance in the LLM responses and ensure reproducibility across \textit{Simulation} runs, we set the temperature parameter of \texttt{GPT-4o} and \texttt{GPT-4o mini} to zero. This deterministic setting allows the same prompt to consistently produce identical outputs, reducing the need for multiple \textit{Simulation} runs. The implemented framework is available on GitHub\footnote{\label{github-link}\url{https://github.com/AISL-at-Imperial-College-London/fault-handling-agentic-llms-for-controlled-operations}}.

\begin{algorithm}[ht]
\footnotesize
\caption{LLM-guided control loop for mixing module}
\label{alg:llm-loop}
\KwIn{Initial plant state with fault parameters (e.g., clogging)}
\KwOut{Actuator settings, plant trajectories, LLM token usage logs}

Initialize \texttt{plant\_states} and kick off \texttt{RouterFlow}\;
\While{not \texttt{terminate}}{
    \textbf{Monitor:} Update process state from plant simulation\;

    \textbf{Generate Action:} Call \texttt{PlantOperatorCrew} to propose actions based on current state and fault type\;

    \textbf{Simulate:} Apply actions to Digital Twin using \texttt{digital\_twin()}\;

    \textbf{Validate:} Check action and pump power validity using \texttt{validation\_script.py}\;

    \uIf{actions are valid}{
        \textbf{Execute:} Forward actions to real plant model using \texttt{plant()}\;
    }
    \uElseIf{$\texttt{reprompt} < \texttt{max\_itr}$}{
        \textbf{Reprompt:} Generate new suggestions using \texttt{PlantStrategyCrew}\;
    }
    \Else{
        \textbf{Force execution:} Pass current action to plant (fallback)\;
    }

    Log process data (\texttt{CSV}), token usage, and reprompt statistics\;

    \If{\texttt{tank\_B204} reaches target level}{
        \texttt{terminate} $\gets$ \texttt{True}\;
    }
}
\textbf{Export results:} Save \texttt{plant\_op.csv}, \texttt{digital\_twin\_op.csv}, \texttt{llm\_plant\_op.csv}\;
\end{algorithm}

\subsection{Evaluation Metrics} 
To evaluate the framework, we define metrics assessing its reliability in generating corrective actions and its efficiency in terms of reprompts needed, directly addressing the \textbf{RQs}.

\textbf{RQ1}, which targets the ability of LLMs to autonomously manage unforeseen faults while ensuring operational safety, is addressed by evaluating the \textbf{Action Quality} and the \textbf{Efficiency} of closed-loop decision-making. In the presented case study, the desired corrective action is the increase of the pump power of \texttt{Pump\_P101} to compensate for the clogging fault. For evaluation purposes, \textbf{Action Quality} is operationalized through five specific metrics: the number of \emph{Correct Actions}, \emph{Incorrect Valve Actions}, \emph{Incorrect Pump Actions}, \emph{Missed Valve Actions}, and \emph{Missed Pump Actions}. These metrics quantify whether the agent-controlled actuator settings resolve the fault and stabilize the process plant without introducing adverse side effects. In this context, the total number of \emph{Actions} is defined as the sum of \emph{Correct} and \emph{Incorrect Actions}, while the number of \emph{Expected Actions} corresponds to the sum of \emph{Correct} and \emph{Missed Actions}. Additionally, we track \emph{Reprompts}, representing the number of iterations needed to reach a valid corrective control action, as an indicator for \textbf{Efficiency}. Fewer reprompts indicate faster decision-making, whereas higher values reflect increased LLM reasoning effort. In contrast, \textbf{RQ2} explores the type and representation of plant-specific information required to enable effective and reliable control decisions. Our hypothesis, grounded in systems engineering principles, is that function, structure, and behavior lead to effective corrective actions. We compare three prompt formats for the \texttt{<Plant Description>}: (i) a natural language description of the system (\textbf{Text}), (ii) structured OpenModelica code (\textbf{Modelica Code}), and (iii) existing engineering artifacts, such as drawings from the \textbf{State Machine (SM)} and \textbf{P\&ID }provided in vector format. While all formats contain system-level information, they differ in their representation modality. The evaluation aims to analyze how these different formats affect the resulting \textbf{Action Quality}. Additionally, the number of \textit{Tokens} was measured to assess the amount of input required by each representation, providing an indication of potential computational costs.

\subsection{Results} \label{Results}
The results, shown in Tables~\ref{tab:gpt4o} and~\ref{tab:gpt4o-mini}, reveal that the proposed framework successfully produces \textit{Correct Actions} in the majority of cases across all representations. For \texttt{GPT-4o}, the \textbf{Text} input resulted in perfect control performance (15/15) correct actions) without any \textit{Incorrect Actions}, and with only a single required \textit{Reprompt}, indicating excellent loop convergence. The \textbf{SM+P\&ID} format also performed well (14/15 correct), with minimal faults and a moderate number of \textit{Reprompts} (5). The \textbf{Modelica Code} format achieved good results (12/15 correct), though it introduced some \textit{Missed Pump Actions} (3) and a higher \textit{Reprompt} count (6), highlighting that behavioral information in \textbf{Modelica Code }is more difficult for the LLM to interpret reliably. 

\texttt{GPT-4o-mini} exhibited similar trends. The \textbf{SM+P\&ID} format achieved strong performance in terms of \textit{Correct Actions} (13/15) but required more \textit{Reprompts} (9), indicating slightly lower reasoning efficiency. Similarly, the \textbf{Text} input yielded 13/15 correct actions with a moderate number of \textit{Reprompts} (6). Again, the \textbf{Modelica Code} representation showed the weakest performance with multiple \textit{Missed Pump Actions} and \textit{Incorrect Valve Actions} as well as the highest number of \textit{Reprompts} (10). The latter suggests again that the \textbf{Modelica Code} format poses challenges for the LLM in interpreting the embedded behavioral logic, reinforcing the earlier observation. 

\textbf{Token Usage}, also reported in Tables~\ref{tab:gpt4o} and~\ref{tab:gpt4o-mini}, varies substantially across representations. The \textbf{Modelica Code} format leads to the highest token consumption (up to 113K), while the \textbf{Text} format remains most efficient. This confirms that the representation modality not only impacts control performance but also computational cost.
\begin{table}
\small
\centering
\renewcommand{\arraystretch}{1.2}
\begin{tabularx}{\linewidth}{l *{3}{>{\centering\arraybackslash}X}}
\toprule
\textbf{Metrics} & \multicolumn{3}{c}{\textbf{\texttt{<Plant Description>}}} \\
\cmidrule(lr){2-4}
& \textbf{Text} & \textbf{Modelica Code} & \textbf{SM + P\&ID} \\
\midrule\midrule
\textbf{Actions Summary} & & & \\
\hspace{1em}No. of Actions & 15 & 12 & 14 \\
\hspace{1em}No. of Expected Actions & 15 & 15 & 15 \\
\\[-0.8em]
\textbf{Action Quality} & & & \\
\hspace{1em}No. of Correct Actions & 15 & 12 & 14 \\
\hspace{1em}No. of Incorrect Valve Actions & 0 & 0 & 0 \\
\hspace{1em}No. of Incorrect Pump Actions & 0 & 0 & 0 \\
\hspace{1em}No. of Missed Valve Actions & 0 & 0 & 0 \\
\hspace{1em}No. of Missed Pump Actions & 0 & 3 & 1 \\
\\[-0.8em]
\textbf{Efficiency} & & & \\
\hspace{1em}No. of Reprompts & 1 & 6 & 5 \\
\\[-0.8em]
\textbf{Token Usage} & & & \\
\hspace{1em}No. of Tokens (K) & 16.2 & 81.4 & 27.2 \\
\bottomrule
\end{tabularx}
\vspace{0.5em}
\captionof{table}{Performance of GPT-4o across different input representations.}
\label{tab:gpt4o}
\end{table}
\begin{table}
\small
\centering
\renewcommand{\arraystretch}{1.2}
\begin{tabularx}{\linewidth}{l *{3}{>{\centering\arraybackslash}X}}
\toprule
\textbf{Metrics} & \multicolumn{3}{c}{\textbf{\texttt{<Plant Description>}}} \\
\cmidrule(lr){2-4}
& \textbf{Text} & \textbf{Modelica Code} & \textbf{SM + P\&ID} \\
\midrule\midrule
\textbf{Actions Summary} & & & \\
\hspace{1em}No. of Actions & 13 & 14 & 14 \\
\hspace{1em}No. of Expected Actions & 15 & 15 & 15 \\
\\[-0.8em]
\textbf{Action Quality} & & & \\
\hspace{1em}No. of Correct Actions & 13 & 12 & 13 \\
\hspace{1em}No. of Incorrect Valve Actions & 0 & 2 & 1 \\
\hspace{1em}No. of Incorrect Pump Actions & 0 & 0 & 0 \\
\hspace{1em}No. of Missed Valve Actions & 0 & 0 & 0 \\
\hspace{1em}No. of Missed Pump Actions & 2 & 3 & 2 \\
\\[-0.8em]
\textbf{Efficiency} & & & \\
\hspace{1em}No. of Reprompts & 6 & 10 & 9 \\
\\[-0.8em]
\textbf{Token Usage} & & & \\
\hspace{1em}No. of Tokens (K) & 33.9 & 113.0 & 40.5 \\
\bottomrule
\end{tabularx}
\vspace{0.5em}
\captionof{table}{Performance of GPT-4o-mini across different input representations.}
\label{tab:gpt4o-mini}
\end{table}

%% file: 07_Discussion.tex
\section{Discussion} \label{discussion}

Tables~\ref{tab:gpt4o} and~\ref{tab:gpt4o-mini} demonstrate that the proposed methodological framework enables a reliable generation of corrective control actions for fault handling in process plants. Both \texttt{GPT-4o} and \texttt{GPT-4o-mini} achieved a high number of \textit{Correct Actions} across all tested representations for \texttt{<Plant Description>}, particularly for the \textbf{Text} variant (15/15 for \texttt{GPT-4o}, 13/15 for \texttt{GPT-4o-mini}) and the \textbf{SM+P\&ID} format (14/15 for \texttt{GPT-4o}, 13/15 for \texttt{GPT-4o-mini}). Most runs required only a few \textit{Reprompts} (1–6), indicating stable loop convergence and minimal computational overhead. Even with more complex input, such as the \textbf{Modelica Code} format, the framework maintained acceptable performance, although an increase in \textit{Incorrect Actions}, \textit{Missed Actions}, and \textit{No. of Tokens} was observed.  This suggests that interpreting the control logic from \textbf{Modelica code} may be more challenging for the LLMs, possibly due to the comparatively lower availability of Modelica-specific training data in public datasets. In contrast, codebases from more commonly used simulation environments, such as MATLAB/Simulink or Python-based frameworks, might offer more familiar structures for the models and could potentially lead to improved performance. All simulation runs and detailed results are made available in the associated GitHub\footnotemark[\value{footnote}] repository.

These findings indicate that the methodological framework, as outlined in Sec.~\ref{Autonomous Fault Handling}, can be successfully applied to support autonomous fault handling in modular process plants, addressing \textbf{RQ1}. Nonetheless, although the results are acceptable, technical systems such as process plants require very high levels of reliability and safety, meaning that further refinements are necessary to achieve consistently perfect results. Regarding \textbf{RQ2}, the results show that structuring the prompt according to system engineering principles, specifically by representing structure, function, and behavior of the system, contributed to the generation of appropriate corrective actions. As Digital Twins typically maintain such structured information, they offer a valuable basis for integrating lifecycle engineering data into the prompt engineering process within such frameworks.

Nonetheless, despite its promising results, the proposed framework has limitations. First, the study focused on batch production processes. Continuous-time systems exhibit more complex dynamics, which may require additional domain knowledge, more reprompts, and as a result longer latencies. Second, although the results support selective information provision, the limited context window of current LLMs restricts how much structured content about the system can be processed per iteration. RAG may mitigate this but introduces additional latency. Finally, the experimental setup only considered a single module of the overall plant. More complex multi-module setups with a greater number of actuators and sensors are likely to increase the difficulty of fault handling. 

%% file: 08_Summary_and_Outlook.tex
\section{Summary and Outlook} \label{summary and outlook}
To progressively enhance autonomy in the fault handling of process plants and reduce the need for manual intervention, this paper introduced a methodological framework that integrates LLM-based agents with a \textit{Digital Process Plant Twin}. The framework is designed to identify faults, derive suitable corrective control actions, and validate those actions through closed-loop \textit{Simulation} before they are applied to the physical plant. Central to this architecture is an iterative cycle involving an \textit{Action Agent}, the \textit{Simulation}, a \textit{Validation Agent}, and a \textit{Reprompting Agent}, which jointly ensure that proposed actions are not only effective but also safe. To enhance the reasoning capabilities of LLMs in this domain, a tailored prompt engineering was developed based on principles from systems engineering, embedding plant-specific knowledge derived from structural, behavioral, and functional models. Application of the framework to a simulated modular process plant demonstrated that effective corrective control actions could be generated efficiently.

Future work should incorporate more expressive behavioral models, such as differential equation-based descriptions, to better reflect the continuous dynamics of physical systems and broaden simulation-based validation. Integrating RAG can enable LLM agents to access structured plant data or documentation in real time, enhancing contextual reasoning. Complementary sub-symbolic AI methods, such as ML-based anomaly detection, may further strengthen the \textit{Monitoring Agent’s} ability to anticipate faults. To address latency in closed-loop interactions between LLM agents and \textit{Simulation}, iteration times must be reduced. Promising approaches include parallelized simulations, targeted state updates, and surrogate or reduced-order models for faster response estimation.

%% file: references.bbl

%% file: paper.bbl
\begin{thebibliography}{10}
\providecommand{\url}[1]{#1}
\csname url@samestyle\endcsname
\providecommand{\newblock}{\relax}
\providecommand{\bibinfo}[2]{#2}
\providecommand{\BIBentrySTDinterwordspacing}{\spaceskip=0pt\relax}
\providecommand{\BIBentryALTinterwordstretchfactor}{4}
\providecommand{\BIBentryALTinterwordspacing}{\spaceskip=\fontdimen2\font plus
\BIBentryALTinterwordstretchfactor\fontdimen3\font minus \fontdimen4\font\relax}
\providecommand{\BIBforeignlanguage}[2]{{%
\expandafter\ifx\csname l@#1\endcsname\relax
\typeout{** WARNING: IEEEtran.bst: No hyphenation pattern has been}%
\typeout{** loaded for the language `#1'. Using the pattern for}%
\typeout{** the default language instead.}%
\else
\language=\csname l@#1\endcsname
\fi
#2}}
\providecommand{\BIBdecl}{\relax}
\BIBdecl

\bibitem{Markaj.2024}
A.~Markaj, M.~Mercang{\"o}z, and A.~Fay, ``{Design and implementation of an Autonomous Systems Training Environment framework for control algorithm evaluation in autonomous plant operation},'' \emph{{Computers {\&} Chemical Engineering}}, vol. 189, p. 108798, 2024.

\bibitem{Webert.2022}
H.~Webert, T.~D{\"o}{\ss} \emph{et~al.}, ``{Fault Handling in Industry 4.0: Definition, Process and Applications},'' \emph{{Sensors}}, vol.~22, no.~6, p. 2205, 2022.

\bibitem{Manca.2021}
G.~Manca and A.~Fay, ``{Detection of Historical Alarm Subsequences Using Alarm Events and a Coactivation Constraint},'' \emph{{IEEE Access}}, vol.~9, pp. 46\,851--46\,873, 2021.

\bibitem{Westermann.2023}
T.~Westermann, M.~S. Gill, and A.~Fay, ``{Representing Timed Automata and Timing Anomalies of Cyber-Physical Production Systems in Knowledge Graphs},'' in \emph{IECON 2023-49th Annual Conference of the IEEE Industrial Electronics Society}, 2023, pp. 1--7.

\bibitem{Gill.2024}
M.~S. Gill, T.~Westermann \emph{et~al.}, ``{Integrating Ontology Design with the CRISP-DM in the Context of Cyber-Physical Systems Maintenance},'' in \emph{2024 IEEE 29th International Conference on Emerging Technologies and Factory Automation (ETFA)}, 2024, pp. 1--8.

\bibitem{Liu.2023}
Y.~Liu, P.~Ramin \emph{et~al.}, ``{Transforming data into actionable knowledge for fault detection, diagnosis and prognosis in urban wastewater systems with AI techniques: A mini-review},'' \emph{{Process Safety and Environmental Protection}}, vol. 172, pp. 501--512, 2023.

\bibitem{GPT4IAS}
Y.~Xia, M.~Shenoy \emph{et~al.}, ``Towards autonomous system: flexible modular production system enhanced with large language model agents,'' in \emph{2023 IEEE 28th International Conference on Emerging Technologies and Factory Automation (ETFA)}, 2023, pp. 1--8.

\bibitem{Vyas24}
\BIBentryALTinterwordspacing
J.~Vyas and M.~Mercang\"{o}z, ``Autonomous industrial control using an agentic framework with large language models,'' 2024. [Online]. Available: \url{https://arxiv.org/abs/2411.05904}
\BIBentrySTDinterwordspacing

\bibitem{Lewis.22.05.2020}
\BIBentryALTinterwordspacing
P.~Lewis, E.~Perez \emph{et~al.}, ``{Retrieval-Augmented Generation for Knowledge-Intensive NLP Tasks},'' 2020. [Online]. Available: \url{http://arxiv.org/pdf/2005.11401}
\BIBentrySTDinterwordspacing

\bibitem{Hildebrandt.2017}
C.~Hildebrandt, A.~Scholz \emph{et~al.}, ``Semantic modeling for collaboration and cooperation of systems in the production domain,'' in \emph{2017 22nd IEEE International Conference on Emerging Technologies and Factory Automation}.\hskip 1em plus 0.5em minus 0.4em\relax Piscataway, NJ: IEEE, 2017, pp. 1--8.

\bibitem{Piardi.2025}
L.~Piardi, A.~S. de~Oliveira \emph{et~al.}, ``{Collaborative fault tolerance for cyber--physical systems: The detection stage},'' \emph{{Computers in Industry}}, vol. 166, p. 104253, 2025.

\bibitem{Abele.2013}
L.~Abele, M.~Anic \emph{et~al.}, ``{Combining Knowledge Modeling and Machine Learning for Alarm Root Cause Analysis},'' \emph{{IFAC Proceedings Volumes}}, vol.~46, no.~9, pp. 1843--1848, 2013.

\bibitem{Thambirajah.2009}
J.~Thambirajah, L.~Benabbas \emph{et~al.}, ``{Cause-and-effect analysis in chemical processes utilizing XML, plant connectivity and quantitative process history},'' \emph{{Computers {\&} Chemical Engineering}}, vol.~33, no.~2, pp. 503--512, 2009.

\bibitem{Kirchhubel.2020}
D.~Kirchh{\"u}bel, M.~Lind, and O.~Ravn, ``{Combining operations documentation and data to diagnose procedure execution},'' \emph{{Computers {\&} Chemical Engineering}}, vol. 140, p. 106940, 2020.

\bibitem{Tao.2019}
F.~Tao, H.~Zhang \emph{et~al.}, ``{Digital Twin in Industry: State-of-the-Art},'' \emph{IEEE Transactions on Industrial Informatics}, vol.~15, no.~4, pp. 2405--2415, 2019.

\bibitem{AshishVaswani.}
\BIBentryALTinterwordspacing
{Ashish Vaswani}, {Noam Shazeer} \emph{et~al.}, ``{Attention is All you Need},'' 2017. [Online]. Available: \url{https://arxiv.org/abs/1706.03762}
\BIBentrySTDinterwordspacing

\bibitem{Liang.}
J.~Liang, W.~Huang \emph{et~al.}, ``{Code as Policies: Language Model Programs for Embodied Control}.''

\bibitem{Bubeck.22.03.2023}
\BIBentryALTinterwordspacing
S.~Bubeck, V.~Chandrasekaran \emph{et~al.}, ``{Sparks of Artificial General Intelligence: Early experiments with GPT-4},'' 2023. [Online]. Available: \url{http://arxiv.org/pdf/2303.12712}
\BIBentrySTDinterwordspacing

\bibitem{Xia.}
Y.~Xia, N.~Jazdi, and M.~Weyrich, ``{Enhance FMEA with Large Language Models for Assisted Risk Management in Technical Processes and Products},'' in \emph{2024 IEEE 29th International Conference on Emerging Technologies and Factory Automation (ETFA), 2024}, pp. 1--4.

\bibitem{HVACLLM}
\BIBentryALTinterwordspacing
L.~Song, C.~Zhang \emph{et~al.}, ``Pre-trained large language models for industrial control,'' 2023. [Online]. Available: \url{https://arxiv.org/abs/2308.03028}
\BIBentrySTDinterwordspacing

\bibitem{LLM4PLC}
M.~Fakih, R.~Dharmaji \emph{et~al.}, ``Llm4plc: Harnessing large language models for verifiable programming of plcs in industrial control systems,'' in \emph{Proceedings of the 46th International Conference on Software Engineering: Software Engineering in Practice}, ser. ICSE-SEIP ’24.\hskip 1em plus 0.5em minus 0.4em\relax ACM, Apr. 2024, p. 192–203.

\bibitem{Agent4PLC}
\BIBentryALTinterwordspacing
Z.~Liu, R.~Zeng \emph{et~al.}, ``Agents4plc: Automating closed-loop plc code generation and verification in industrial control systems using llm-based agents,'' 2024. [Online]. Available: \url{https://arxiv.org/abs/2410.14209}
\BIBentrySTDinterwordspacing

\bibitem{Cerrada.2007}
M.~Cerrada, J.~Cardillo \emph{et~al.}, ``{Agents-based design for fault management systems in industrial processes},'' \emph{{Computers in Industry}}, vol.~58, no.~4, pp. 313--328, 2007.

\bibitem{Cummins.15.01.2024}
L.~Cummins, A.~Sommers \emph{et~al.}, ``{Explainable Predictive Maintenance: A Survey of Current Methods, Challenges and Opportunities},'' \emph{{IEEE Access}}, vol.~12, pp. 57\,574 -- 57\,602, 2024.

\bibitem{J.Ehrhardt.2022}
{J. Ehrhardt}, {M. Ramonat} \emph{et~al.}, ``An {AI} benchmark for diagnosis, reconfiguration {\&} planning,'' in \emph{2022 IEEE 27th International Conference on Emerging Technologies and Factory Automation (ETFA)}, 2022, pp. 1--8.

\end{thebibliography}
